\documentclass{article}
\pdfoutput=1
\usepackage{arxiv}
\usepackage{nicefrac}       
\usepackage{microtype}      
\usepackage{booktabs}       
\usepackage[numbers]{natbib}
\usepackage{doi}
\usepackage{adjustbox}
\usepackage{float}
\usepackage[T1]{fontenc}
\usepackage{graphicx}
\usepackage[utf8]{inputenc}
\DeclareUnicodeCharacter{FF1A}{:}  
\usepackage{multicol}
\usepackage{multirow} 
\title{SciGraphQA: A Large-Scale Synthetic Multi-Turn Question-Answering Dataset for Scientific Graphs}

\author{
    Shengzhi Li \\
    alex.shengzhi@gmail.com \\
   \And
    Nima Tajbakhsh \\
    n.tajbakhsh63@gmail.com \\
}

\renewcommand{\headeright}{}
\renewcommand{\shorttitle}{SciGraphQA: A Large-Scale Synthetic Multi-Turn Question-Answering Dataset for Scientific Graphs}

\begin{document}
\maketitle

\begin{abstract}

In this work, we present SciGraphQA, a synthetic multi-turn question-answer dataset related to academic graphs. SciGraphQA is 13
times larger than ChartVQA, the previously largest chart-visual question-answering dataset. It is also
the largest open-sourced chart VQA dataset with non-synthetic charts. To build our dataset, we selected 290,000 Computer Science or Machine Learning ArXiv papers published between 2010 and 2020, and then used Palm-2 to generate 295K samples of open-vocabulary multi-turn question-answering dialogues about the graphs. As context, we provided the text-only Palm-2 with paper title, abstract, paragraph mentioning the graph, and rich text contextual data from the graph itself, obtaining dialogues with an average 2.23 question-answer turns for each graph. We asked GPT-4 to assess the matching quality of our question-answer turns given the paper's context, obtaining an average rating of 8.7/10 on our 3K test set. 

We evaluated the 0-shot capability of the most popular MLLM models such as LLaVa, mPLUGowl, BLIP-2, and openFlamingo's on our dataset, finding LLaVA-13B being the most performant with a CIDEr score of 0.08. We further enriched the question prompts for LLAVA by including the serialized data tables extracted from the graphs using the DePlot model, boosting LLaVA's 0-shot CIDEr to 0.15. To verify the validity of our dataset, we also fine-tuned LLaVa using our dataset, reaching a substantially higher CIDEr score of 0.26. We anticipate further accuracy improvement by including segmentation mask tokens and leveraging larger LLM backbones coupled with emergent prompting techniques. Our code and data are open-sourced at \hyperlink{https://github.com/findalexli/SciGraphQA}{Github} and \hyperlink{https://huggingface.co/datasets/alexshengzhili/SciGraphQA-295K-train}{HuggingFace dataset}.

\end{abstract}


\section{Introduction}

Scientific literature is a rich repository, teeming with detailed graphs, diagrams, and figures that convey intricate data narratives. The task of comprehending, describing, and interpreting these graphs requires significant time and effort from students, data science practitioners, and researchers. Unlike natural images, which are interpreted based on objects, their relative positions, and interactions, scientific graphs carry distinct semantic meanings, illustrated through abstract components such as trend lines and color-coded legends. Typically, readers interpret these graphs in the context of the paper's abstract and pertinent paragraphs. In a live presentation scenario, the author would initiate the context, engaging the audience in a multi-turn, question-answering dialogue.

Inspired by this interactive process, we have developed SciGraphQA, a comprehensive, multi-turn question-answering dataset derived from scientific literature and graphs. Leveraging a commercial large-language model, we have generated question-answering dialogues, providing it with the figure caption, paper title, abstract, and the initial paragraph referencing the figure. Our objective with this dataset is to establish a scientific question-answering benchmark for multi-modal large language models (MLLMs). In this paper, we evaluate the performance of current MLLMs and introduce a baseline model, SciGraphQA-baseline, which has been trained on the SciGraphQA training set.

\section{Related Work}

\subsection{Graph Visual Answering datasets}
\begin{table}[!htbp]
\renewcommand{\arraystretch}{1.3}
\begin{adjustbox}{max width=\textwidth}
\begin{tabular}{p{2.33cm}p{2.33cm}p{2.33cm}p{2.33cm}p{2.33cm}p{2.33cm}p{2.99cm}p{2.33cm}p{2.33cm}p{2.33cm}p{2.33cm}p{2.33cm}p{2.33cm}p{2.99cm}}
\hline
\multicolumn{1}{|p{2.33cm}}{Datasets} & 
\multicolumn{1}{|p{2.33cm}}{Figure Count} & 
\multicolumn{1}{|p{2.33cm}}{Data/Chart generation process} & 
\multicolumn{1}{|p{2.33cm}}{Question-Answer pair count} & 
\multicolumn{1}{|p{2.33cm}}{Question generation} & 
\multicolumn{1}{|p{2.33cm}}{Answer Type} & 
\multicolumn{1}{|p{2.99cm}|}{$\#$ Plot types} \\ 
\hline
\multicolumn{1}{|p{2.33cm}}{FigureQA \cite{kahou2017figureqa}} & 
\multicolumn{1}{|p{2.33cm}}{180K} & 
\multicolumn{1}{|p{2.33cm}}{Synthetic data and charts} & 
\multicolumn{1}{|p{2.33cm}}{2.3M } & 
\multicolumn{1}{|p{2.33cm}}{From 15 templates} & 
\multicolumn{1}{|p{2.33cm}}{Fixed voca. } & 
\multicolumn{1}{|p{2.99cm}|}{4} \\ 
\hline
\multicolumn{1}{|p{2.33cm}}{DVQA \cite{kafle2018}} & 
\multicolumn{1}{|p{2.33cm}}{300K} & 
\multicolumn{1}{|p{2.33cm}}{Synthetic  data and synthetic charts} & 
\multicolumn{1}{|p{2.33cm}}{3.4M} & 
\multicolumn{1}{|p{2.33cm}}{From 26 templates} & 
\multicolumn{1}{|p{2.33cm}}{Fixed vocab. } & 
\multicolumn{1}{|p{2.99cm}|}{1} \\ 
\hline
\multicolumn{1}{|p{2.33cm}}{PlotQA \cite{methani2020plotqa}} & 
\multicolumn{1}{|p{2.33cm}}{224K} & 
\multicolumn{1}{|p{2.33cm}}{Real-world data and synthetic charts} & 
\multicolumn{1}{|p{2.33cm}}{28M} & 
\multicolumn{1}{|p{2.33cm}}{From 76 templates } & 
\multicolumn{1}{|p{2.33cm}}{Mix of fixed and open vocabulary answers. } & 
\multicolumn{1}{|p{2.99cm}|}{3 \newline
} \\ 

\hline
\multicolumn{1}{|p{2.33cm}}{ChartQA \cite{masry2022chartqa}} & 
\multicolumn{1}{|p{2.33cm}}{21.9K (4.8K human and 17.1K generated) } & 
\multicolumn{1}{|p{2.33cm}}{Real-world charts from a web crawl} & 
\multicolumn{1}{|p{2.33cm}}{32.7K (9.6K human and 23.1k generated)} & 
\multicolumn{1}{|p{2.33cm}}{Human/Machine generated} & 
\multicolumn{1}{|p{2.33cm}}{Open Vocabulary} & 
\multicolumn{1}{|p{2.99cm}|}{Unbounded (real-world charts)} \\ 
\hline
\multicolumn{1}{|p{2.33cm}}{SciGraphQA (ours)} & 
\multicolumn{1}{|p{2.33cm}}{295K} & 
\multicolumn{1}{|p{2.33cm}}{Real-world academic graphs } & 
\multicolumn{1}{|p{2.33cm}}{657K} & 
\multicolumn{1}{|p{2.33cm}}{Machine Generated with Palm with additional textual context} & 
\multicolumn{1}{|p{2.33cm}}{Open Vocabulary} & 
\multicolumn{1}{|p{2.99cm}|}{Unbounded (real-world charts)} \\ 
\hline
\end{tabular}
\end{adjustbox}
\caption{Related open datasets on graph visual questioning.}
\label{tab:related_open_datasets_graph_visual}\end{table}
Table \ref{tab:related_open_datasets_graph_visual} presented a series of open-source datasets pertinent to graph/chart visual question answering. Earlier datasets such as FigureQA \cite{kahou2017figureqa}, DVQA \cite{kafle2018}, and LEAF-QA \cite{chaudhry2019} primarily relied on synthetic data, with template-generated questions and answers selected from a fixed vocabulary. In contrast, PlotQA \cite{methani2020plotqa} introduced open-vocabulary questions. Built on ground truth data tables crawled online, PlotQA programmatically generated charts, and questions answers, but the limitations stem from its synthetic nature post data crawl,  restricted chart types: bar plots, line plots, and scatter plots and restricted question variety \cite{methani2020plotqa}

More recently, ChartQA capitalized on real-world, web-crawled charts to develop its visual question-answering datasets, supplemented by human annotators. In order to scale up, ChartQA's authors fine-tuned a T5 model to generate 2/3 of their questions and answers derived from human-written chart summaries \cite{masry2022chartqa}. 

The SciGraphQA dataset is a large-scale captioning dataset that utilizes over 290,000 Arxiv papers focused on Computer Science (cs) and Machine Learning (stat.ML) \cite{hsu2021scicap}. PDFFigures 2.0 was used to extract figures. Out of 2.1 million figures extracted, the top categories are tables (23.6$\%$), graphs (19.2$\%$), and flowcharts (8.5$\%$). Graphs were chosen as the focus for caption prediction. Despite limited success in predicting captions based on images (BLEU-4 $\sim$ 0.02), SciGraphQA provided valuable insights by demonstrating that text normalization did not enhance model performance. Its successor, SciGraphQA+, added more context to image captioning by incorporating the first paragraph that referenced the figure and OCR-extracted text, yielding an increase in BLEU-4 (0.064).

The SciGraphQA/SciGraphQA+ dataset, however, has inherent limitations when framing the problem as a caption prediction task. The ground truth captions may not always be helpful or explicit enough. One-third of captions in SciGraphQA are single-sentence, many of which are incomplete sentences (e.g., "Vary the number of objects") or consist solely of single/multiple nouns (e.g., 'IIPS', 'Average Travel Time'). For a user's perspective, such as a scientific writer, automatic captioning might not reduce writing time, especially if the system requires the writer to supply context, as in the case with SciGraphQA+. If a full-bodied description is necessary for caption generation to work, we posit that users would prefer to write a short caption themselves.

 We believed a dialog experience without needing lengthy textual input offers a more natural flow of user experience. The human user would only need to upload an image, which could range from a Tableau screenshot by a business analyst, a matplotlib plot from a data scientist, or a multi-colored diagram from a color-blind individual seeking to understand what a red line represents. The human user can guide the conversation by providing feedback or additional context. The questions could be conceptual, graphical or complex reasoning-related. 

With this in mind, we introduced the SciGraphQA dataset with multi-turn question-answer conversations on scientific graphs. The questions and corresponding answers were generated from Palm-2, using SciGraphQA figures, captions, context, and paper metadata. SciGraphQA distinguishes itself from previous datasets by using real-world academic graphs rather than synthetic data or charts. The questions and answers inherently support open vocabulary. By leveraging SciGraphQA and Palm2, we significantly scaled up the ChartQA-style real-world Chart-vqa dataset to 295K question-answer pairs from 21.9 (13X) number of samples as shown in Table \ref{tab:related_open_datasets_graph_visual}. 

\subsection{Chart VQA expert systems}\label{chartvqaexpertsystems}
Previously, when datasets only contained fixed-vocabulary questions and answers, the standard approach was to construct classification-based VQA models such as STL-CQA \cite{singh2020stl}, LEAF-Net \cite{chaudhry2019}, DQVQ-baseline \cite{kafle2018}, and VisionTAPAS \cite{masry2022chartqa}. These encoder-only models encoded image features and textual questions separately, combining them later using attention blocks \cite{chaudhry2019}; \cite{singh2020stl} \cite{kafle2018}.

DePlot/Matcha is a lineage of image-encoder-text-decoder models distinct from the aforementioned encoder-only architectures. These models are trained to 'derender' the chart, meaning they extract text and datapoints needed to render the chart. They are based on the Pix2Struct model, which was pre-trained on website visual understanding (from screenshot to HTML code) \cite{lee2023pix2struct}. Matcha fine-tuned Pix2Struct on various datasets, such as Github IPython notebooks for [chart > code], a mix of PlotQA, web-crawled data, and Wikipedia tables for [chart > table], and math reasoning datasets for [image > answer] \cite{liu2023}. 

DePlot, in turn, was further fine-tuned on [chart > table] datasets, including ChartQA, to specialize in converting charts into linearized tables with titles, legends, and interpolated data point values \cite{liu2022deplot}. The output of Deplot is a text string data table that can be fed into such large autoregressive decoders as GPT-3 or Palm. The combination of Deplot and LLM is currently a state-of-the-art benchmark-tuned model for the ChartQA and PlotQA datasets, surpassing VisionTAPAS with OCR input without the need for an external OCR system \cite{liu2023} \cite{liu2022deplot}.

We proposed to leverage Deplot as an external model to enhance the MLLMs' capabilities with regards to charts. Even the most powerful large multi-modal models cannot match expert OCR models by a significant margin \cite{liu2023hidden}. Zero-shot text recognition results range from 37-61\% compared to the supervised SOTA of 85\%. This deficiency is particularly detrimental to science graphs, where text is key to understanding them. Deplot not only offers OCR capabilities but also can interpolate data point values. This choice to leverage an external system is similar to previous pipelines that incorporated external OCR systems into their input, such as Layout-LM \cite{xu2020layoutlm} and ChartBert \cite{akhtar2023reading}.

\subsection{Multi-modal large language models (MLLM)}

In the domain of natural image understanding and multi-modal generative models, there has been rapid development of vision-language foundational models. These models are similar to the Deplot model discussed in the previous section, as they are image-encoder-text-decoder models. However, the primary distinction lies in the use of pre-trained image encoders, such as CLIP, coupled with a pre-trained large language model backbone, which is typically frozen or partially frozen during training. To bridge the two modalities, MLLMs are trained with 'adapters' that translate image patch embeddings into token embeddings that can be understood by a large language model \cite{yin2023survey}. Notably, these models were trained on natural images rather than text-heavy charts and figures.

The BLIP-2 model employed a Querying Transformer (Q-Former) to bridge the modality gap between vision and language models. By employing learnable query vectors, BLIP-2 is capable of extracting visual representations that are relatable to the language model, \cite{li2023}.

 DeepMind's Flamingo as well as its open-source equivalent OpenFlamingo, designed gated cross-attention-dense layers within frozen pre-trained LLMs and perceiver resampler to support fixed-sized input with variable in-context learning examples. By sourcing keys and values from visual features and queries from language inputs, it efficiently conditions LLMs on visual input. This structure is highly efficient for in-context learnings \cite{alayrac2022flamingo}, \cite{anas2023}).

 LLaVA employed self-instruction with GPT-4 to generate a high-quality multi-modal language-image instruction-following dataset. After training a projection layer with image-text pairs from the CC3M dataset as a feature alignment stage, the language model and projection layer were jointly trained for instruction-tuning. This enabled LLaVA to perform various visual understanding and reasoning tasks  \cite{liu2023}. LLaVA used a frozen CLIP Vit-L/14 model. 

mPLUG-Owl followed a two-stage training process similar to LLaVA, with the disinction of unfrozeing the visual encoder during the first stage and the use of fixed learnable tokens to summarize visual information via cross-attention with image patch features to connect image and text modalities rather an a linear projector. \cite{ye2023}. The first-stage training was done on a significantly larger data mixture that includes LAION-400M \cite{schuhmann2021laion}, COYO-700M \cite{kakaobrain2022coyo-700m}, Conceptual Captions \cite{sharma2018conceptual}, and MSCOCO \cite{lin2014microsoft}. The second-stage instruction dataset includes Vicuna's sharegpt \cite{vicuna2023} and LLaVA (multi-modal) \cite{liu2023}

In this study, we proposed to evaluate the abilities of the MLLMs on our SciGraphQA datasets. We performed zero-shot evaluations to assess their performance in this out-of-distribution domain of scientific graphs. Furthermore, we fine-tuned a baseline MLLM using the SciGraphQA training set.
\section{Methodology}
\subsection{Dataset Construction}
\begin{figure}[!htbp]
\centering
\includegraphics[width=16.504cm,height=16.252cm]{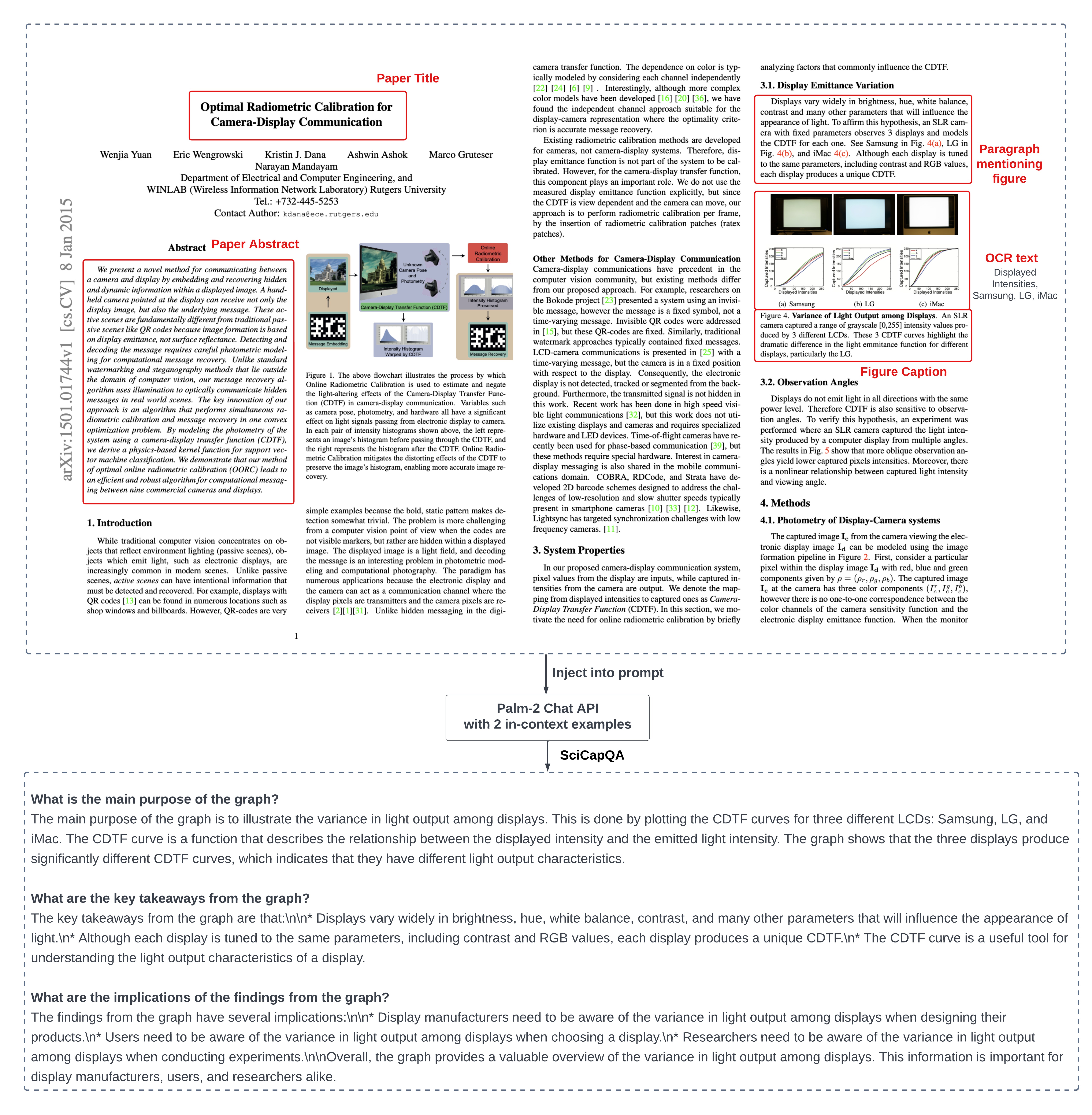}
\caption{Illustration of multi-turn dialogue generation process. For higher quality dialogues, we use comprehensive textual context together with in-context learning when prompting Palm-2.}
\label{fig:questionanswer_generation_process_flow_palm2}
\end{figure}

We expanded the SciCap+ dataset to create the SciGraphQA dataset by generating multi-turn conversations centered around the understanding and elaboration of the content presented in academic graphs. This was achieved by collecting comprehensive textual context related to the graph, which includes the title, abstract, original captions, the first paragraphs mentioning the figure, and OCR text. The SciCap+ dataset is an image-caption dataset derived from scientific figures and captions; it includes additional text derived from the Google OCR API and the paragraphs referenced in the paper \cite{yang2023scicap+}. As shown in figure \ref{fig:questionanswer_generation_process_flow_palm2}, we combined figures, captions, paragraphs, and text-from-OCR from the SciCap+ dataset with the corresponding paper titles and abstracts, and subsequently prompted the PALM2 Chat API to generate multi-turn conversations. We provided this API with two in-context examples generated by GPT-4, as seen in the Appendix \ref{appendix}. In the prompt, we specified that the conversation should focus primarily on the graph, rather than the broader context. For example, we posed questions such as, "Considering the graph structure, can you elaborate on how the lines, axes, and legend provide a comprehensive perspective of the interplay between source variance and mean target risk?" After a filtering process, we were able to curate 295K high-quality, multi-turn data entries. 

As depicted in Figure \ref{fig:datasetexamples}, our approach diverged from LLaVA's methodology of using a fixed set of questions to prompt GPT-4 \cite{liu2023}. We observed a greater diversity in our question set, owing to their generation process conditioned on text context. Some questions were broad, inquiring about textual aspects (e.g., "What are the limitations of this graph?") or components of the composition (e.g., "What does this axis signify?"). On the other hand, several questions delved into specific concepts, mirroring natural conversations by incorporating follow-up queries based on the previous question-answer exchange. This approach seeks to capture the dynamics of organic dialogue, introducing a layer of complexity not present in the fixed-question methodology.

\begin{figure}[!htbp]
\centering
\includegraphics[width=15.3cm,height=5.4cm]{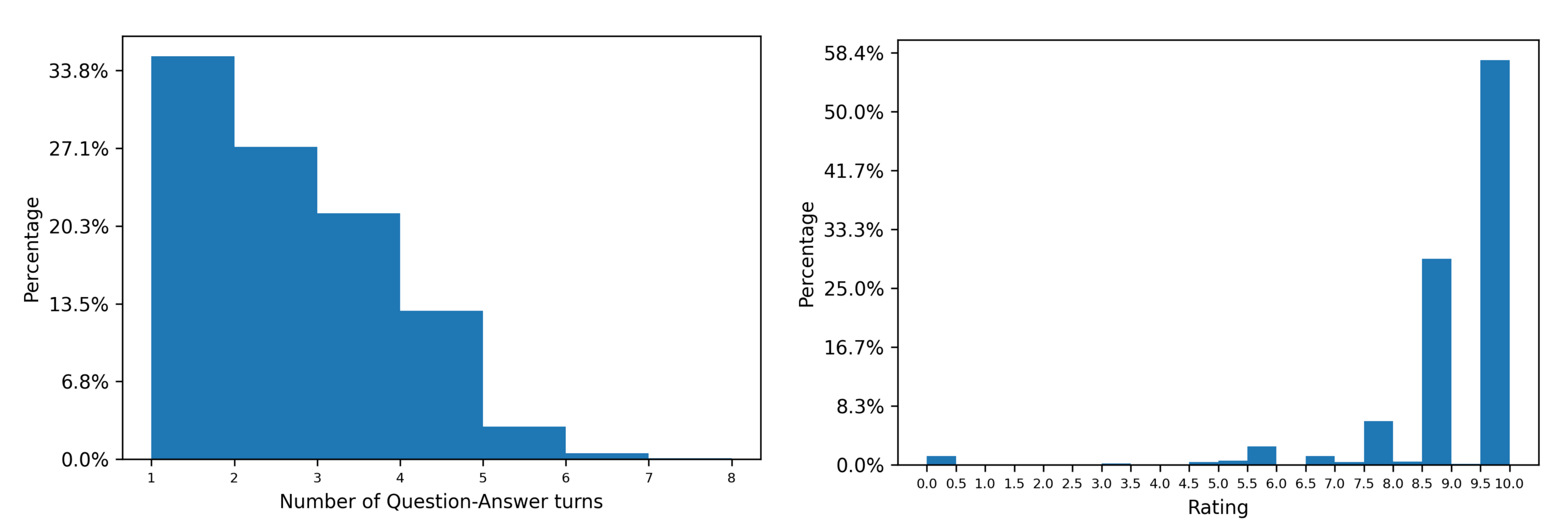}
\caption{(left) distribution of the number of question-answer turns in our SciGraphQA dataset. (right) distribution of GPT-4 ratings (0--10) when GPT-4 was used as a judge to measure the matching of questions and answers from a 3k subset of the the SciGraphQA dataset.}
\label{fig:gpt4_rating_distribution_histogram_gpt4}
\end{figure}

\begin{figure}[!htbp]
\centering
\includegraphics[width=16.504cm,height=21.45cm]{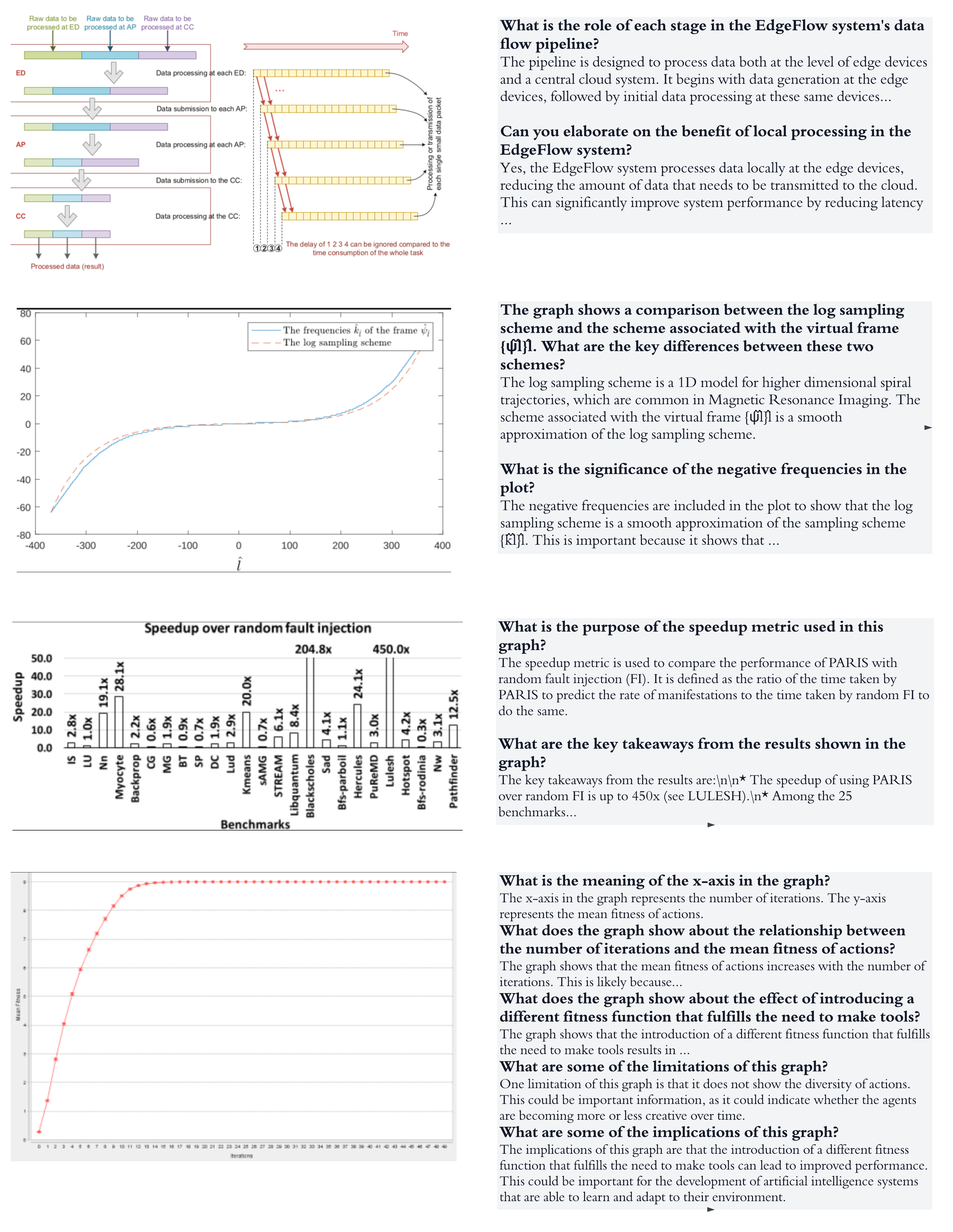}
\caption{Examples from our SciGraphQA dataset where questions and answers are both generated using a commercial LLM rather than being selected from a fixed, limited template pool. Note how the questions are specific to the graphs and often have a conversational nature, asking to elaborate on a concept mentioned in previous answer. For brevity, some answers are truncated, denoted by ``...`` at the end.}
\label{fig:datasetexamples}
\end{figure}

Our initial Palm-2 output contained 350K samples. Manual inspection showed that a portion of the questions are not graph-related but follow-up on a textual context in previous answers. We would like to focus on graph-related questions to evaluate the graph-understanding capability of multi-modal large language models. Hence we developed a heuristic rule to filter out questions that do not contain any keywords in the following list: ['graph', 'diagram', 'figure', 'chart', 'axis', 'plot', 'table', 'image', 'visual', 'illustrat'].  Admittedly, in naturally occurring conversations around scientific graphs, humans ask follow-up questions on abstract concepts as well. We will release the unfiltered question-answers in addition to filtered question-answers. 

After filtering out low-quality QA turns, we curated 295K multi-turn data with 59.1 Million tokens, counted using a byte-wise BPE tokenizer. The average turn is 2.23 while there are 111K samples with 3 or more turns, and we plotted the number of question-answer turns in figure \ref{fig:gpt4_rating_distribution_histogram_gpt4}. On average, we have 143 characters in questions and 775 characters in answers, or 39 tokens in questions and 164 tokens in answers. The token per sample is 199$\pm$98 (mean $\pm$ std). The statistics are similar to LLaVA whose instruction tuning dataset has 109 million tokens with an average of 276 tokens per sample.


To check the quality of our question-answer matching, we used GPT-4 to rate the responses in the full 3K test set using GPT-4. GPT-4 was prompted with the textual context (paragraph, caption, abstract), question, and reference answer and asked to rate the quality and helpfulness of the reference answer to the given questions. As shown in the figure \ref{fig:gpt4_rating_distribution_histogram_gpt4}, we observed that the distribution is heavily skewed towards 10, with 86\% of the answers being rated 8.5/10 or higher. On the low end, there are 6.4\% of ratings of 7/10 or lower, reflecting imperfections in our question-answering generation process. There are 1.2\% of ratings $\leq$ 1/10, meaning the answer does not address the question. We concluded that overall, the GPT-4 rating assessment confirmed the validity of our question-answer generation and filtering process, with the majority of answers accurately answering the question.

\subsection{Zero-shot evaluation}
\label{zero-shot-method}
We were interested in examining the capabilities of existing Multimodal Large Language Models (MLLMs), originally trained with natural images typically sourced from the CoCo or LAION datasets \cite{yin2023survey}, in comprehending and answering questions related to scientific graphs. We used the checkpoints of the candidate models in a zero-shot evaluation setting where they received only the image and question. The test set consisted of 3K pairs of images, questions, and answers from SciGraphQA. We calculated NLP metrics, including BLEU-4, CIDEr, and ROUGE, between the predicted answers and the ground truth for the test set. 

\begin{figure}[!htbp]\label{fig:deplot}
\centering
\includegraphics[width=13.76cm,height=4.31cm]{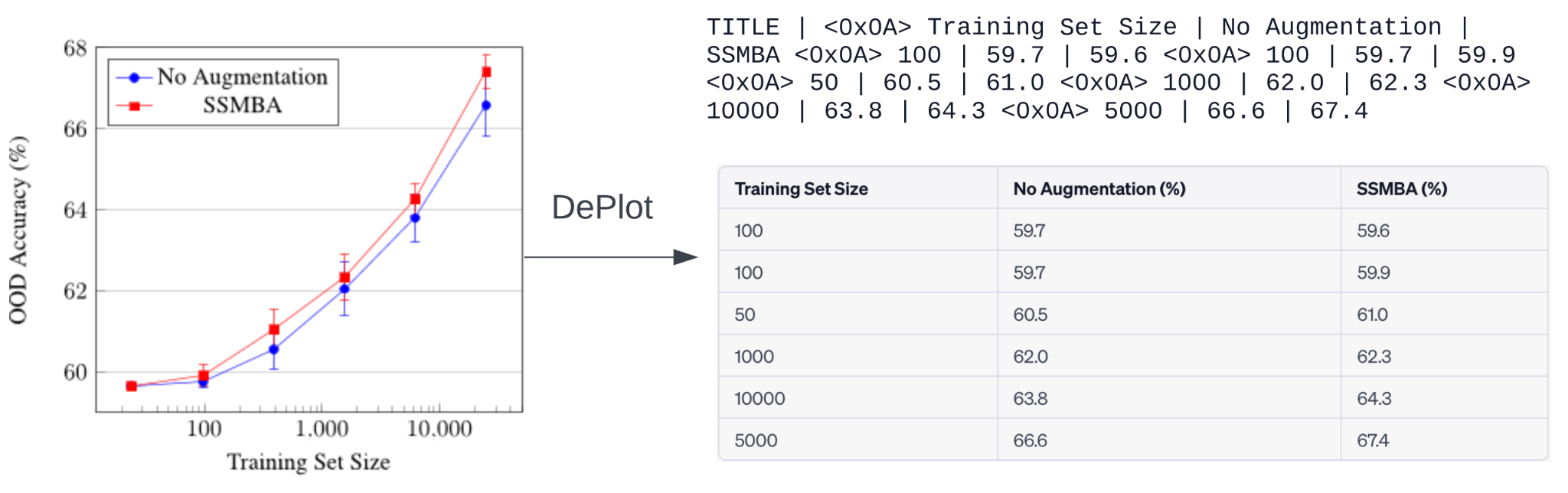}
\caption{Illustration of how the DePlot model extracts the data table from a figure followed by serializing the table into a string.  \textless{}0x0A\textgreater{} is a newline token in hexadecimal format. Note how the data table contains the interpolated values along the x-axis and y-axis for both legends. We used the DePlot's output string as a prompt augmentation technique in our work.}
\end{figure}

Inspired by DePlot\cite{liu2022deplot}, We further augmented the question prompts by prepending them with the data tables extracted from the graphs using the DePlot model, which interpolates the axis legends and numerical values from the given image, as shown in Figure \ref{fig:deplot}. The authors of DePlot showed that GPT 3.5, when prompted with only the tabular data, was able to respond to questions related to a given chart, {\bf underscoring the effectiveness of the tabular data in enhancing the zero-shot capability of LLMs.}
\subsection{Fine-tuning}\label{finetuningmethod}
We further validate our dataset by fine tuning LLaVA-13B, which demonstrated the highest accuracy among the models in the 0-shot evaluation setting. For this purpose, we utilized the LLaVA-13B checkpoint, released after feature alignment and multi-modal instruction fine-tuning \cite{liu2023}. We evaluated the fine-tuned models using the 0-shot evaluation method introduced in the previous section.
We conducted a two-step fine-tuning process for LLaVA-13B on the SciGraphQA dataset. First, we fine-tuned it for 1 epoch on the full dataset, which consisted of 295K samples. Next, we performed fine-tuning on a 30K subset of SciGraphQA, where the question prompts were augmented with the data tables extracted using DePlot from the graphs.

We used a LLaVA checkpoint with LLAMA-2-13B-chat backbone as base model for finetuning. We employeed a cosine learning rate schedule with a warmup ratio of 3\% and a learning rate of \(5 \times 10^{-6}\). This learning rate was deliberately lower than the original \(2 \times 10^{-5}\) used for the instruction tuning of LLaVA and also lower than the learning rate of \(3 \times 10^{-4}\) used for LLAMA-2-13B \cite{liu2023}, \cite{touvron2023}. We made this decision based on early experiments, which indicated that end-to-end fine-tuning with higher learning rates, like \(2 \times 10^{-5}\), resulted in degraded performance. Thus, it underscored the significance of using an appropriate learning rate for successful fine-tuning.

We utilized LoRa to enable us to train small-scale adapters on the LLaMa-2-chat-13B backbone \cite{touvron2023}. We used a LoRa rank of 64 and a LoRa dropout rate of 0.05. Together with DeepSpeed Zero-2 distributed training, this setup allowed us to train with 4 A100-80Gb GPUs on Azure with a per-device batch size of 16 and a gradient accumulation step of 2, resulting in a effective global batch size of 128. 

\section{Results}

\begin{table}[!htbp]\label{mainresultstable}
\renewcommand{\arraystretch}{1.3}
\begin{adjustbox}{max width=\textwidth}
\begin{tabular}{p{4.13cm}p{2.48cm}p{2.48cm}p{2.48cm}p{2.48cm}p{2.48cm}p{4.13cm}p{2.48cm}p{2.48cm}p{2.48cm}p{2.48cm}p{2.48cm}}
\hline
\multicolumn{1}{|p{4.13cm}}{Model Name} & 
\multicolumn{1}{|p{2.48cm}}{Finetuned on SciGraphQA?} & 
\multicolumn{1}{|p{2.48cm}}{Prompt Augmented with extracted data-table} & 
\multicolumn{1}{|p{2.48cm}}{CIDEr} & 
\multicolumn{1}{|p{2.48cm}}{BLEU(4)} & 
\multicolumn{1}{|p{2.48cm}|}{ROUGE} \\ 
\hline
\multicolumn{1}{|p{4.13cm}}{BLIP2-2.7B} & 
\multicolumn{1}{|p{2.48cm}}{No} & 
\multicolumn{1}{|p{2.48cm}}{No} & 
\multicolumn{1}{|p{2.48cm}}{0.007} & 
\multicolumn{1}{|p{2.48cm}}{0.003} & 
\multicolumn{1}{|p{2.48cm}|}{0.1} \\ 
\hline
\multicolumn{1}{|p{4.13cm}}{DePlot+mPLUG-owl-7B} & 
\multicolumn{1}{|p{2.48cm}}{No} & 
\multicolumn{1}{|p{2.48cm}}{Yes} & 
\multicolumn{1}{|p{2.48cm}}{0.037} & 
\multicolumn{1}{|p{2.48cm}}{0.058} & 
\multicolumn{1}{|p{2.48cm}|}{0.22} \\ 
\hline
\multicolumn{1}{|p{4.13cm}}{mPLUG-owl-7B} & 
\multicolumn{1}{|p{2.48cm}}{No} & 
\multicolumn{1}{|p{2.48cm}}{No} & 
\multicolumn{1}{|p{2.48cm}}{0.04} & 
\multicolumn{1}{|p{2.48cm}}{0.062} & 
\multicolumn{1}{|p{2.48cm}|}{0.22} \\ 
\hline
\multicolumn{1}{|p{4.13cm}}{LLaVa-7B} & 
\multicolumn{1}{|p{2.48cm}}{No} & 
\multicolumn{1}{|p{2.48cm}}{No} & 
\multicolumn{1}{|p{2.48cm}}{0.048} & 
\multicolumn{1}{|p{2.48cm}}{0.07} & 
\multicolumn{1}{|p{2.48cm}|}{0.18} \\ 
\hline
\multicolumn{1}{|p{4.13cm}}{LLaVa-13B} & 
\multicolumn{1}{|p{2.48cm}}{No} & 
\multicolumn{1}{|p{2.48cm}}{No} & 
\multicolumn{1}{|p{2.48cm}}{0.08} & 
\multicolumn{1}{|p{2.48cm}}{0.07} & 
\multicolumn{1}{|p{2.48cm}|}{0.23} \\ 
\hline
\multicolumn{1}{|p{4.13cm}}{OpenFlamingo v2-7B} & 
\multicolumn{1}{|p{2.48cm}}{No} & 
\multicolumn{1}{|p{2.48cm}}{No} & 
\multicolumn{1}{|p{2.48cm}}{0.12} & 
\multicolumn{1}{|p{2.48cm}}{0.081} & 
\multicolumn{1}{|p{2.48cm}|}{0.22} \\ 
\hline
\multicolumn{1}{|p{4.13cm}}{DePlot+GPT-3} & 
\multicolumn{1}{|p{2.48cm}}{No} & 
\multicolumn{1}{|p{2.48cm}}{Yes} & 
\multicolumn{1}{|p{2.48cm}}{0.13} & 
\multicolumn{1}{|p{2.48cm}}{0.098} & 
\multicolumn{1}{|p{2.48cm}|}{0.226} \\ 
\hline
\multicolumn{1}{|p{4.13cm}}{DePlot+LLaVa-13B} & 
\multicolumn{1}{|p{2.48cm}}{No} & 
\multicolumn{1}{|p{2.48cm}}{Yes} & 
\multicolumn{1}{|p{2.48cm}}{0.153} & 
\multicolumn{1}{|p{2.48cm}}{0.106} & 
\multicolumn{1}{|p{2.48cm}|}{0.273} \\ 
\hline
\multicolumn{1}{|p{4.13cm}}{DePlot+SciGrahQA-baseline} & 
\multicolumn{1}{|p{2.48cm}}{Yes} & 
\multicolumn{1}{|p{2.48cm}}{Yes} & 
\multicolumn{1}{|p{2.48cm}}{\textbf{0.268}} & 
\multicolumn{1}{|p{2.48cm}}{\textbf{0.123}} & 
\multicolumn{1}{|p{2.48cm}|}{\textbf{0.31}} \\ 
\hline
\end{tabular}
\end{adjustbox}
\caption{Evaluation of large multi-modal models using CIDEr, BLEU-4, and ROUGE metrics on the 3K test set of our SciGraphQA dataset. The term "DePlot + {Model}" denotes prompt-augmentation using DePlot-extracted data tables. The SciGraphQA-baseline is a LLaVA-13B model fine-tuned on SciGraphQA's  training set.}
\label{tab:evaluation_large_multimodal_models_performance}\end{table}
We evaluated the performance of multiple models on our dataset in zero-shot and fine-tuning settings. We set aside a test set of 3K samples and computed NLP metrics including BLEU-4, ROUGE, and CIDEr. As seen in Table~\ref{mainresultstable}, performance of large multi-modal models increased proportionally to the size of their Language-to-Language Model (LLM) backbones. For instance, BLIP2, with its comparatively smaller 2.7B backbone, showed the least accuracy among the models studied. Conversely, LLaVA-13B, equipped with a significantly larger backbone, demonstrate superior performance.

Interestingly, within the 7B range, distinct performances emerged from models like OpenFlamingo (Moasic-7B), LLaVA-7B (Vicuna-7B), and mPLUG-owl (Llama-7B). This disparity in performance could be tied back to their unique multi-modal adapter designs—ranging from linear projectors to a combination of perceived resamplers and cross attention—and the nature of their respective training datasets. The findings point towards a nuanced understanding that simply scaling up model sizes might not be a one-size-fits-all solution, with the design of adapters and choice of training data playing an equally important role.


Despite trailing other 7B MLLMs models, OpenFlamingo's underperformance can be attributed to its unique optimization for in-context learning through a retrieval-based in-context example selection (RICE) scheme, as noted by \cite{yang2022empirical}. This approach retrieves images similar to the query image based on image encodings, and subsequently incorporates top-N (image, text) examples into the prompt, as described in \cite{alayrac2022flamingo}. However, within the constraints of our research, we were unable to replicate this specific methodology. Instead, we augmented the prompts by randomly selecting 3, 6, and 9 examples from the training dataset as in-context examples. Preliminary evaluations of our randomly selected few-shot prompting, however, did not reveal any significant performance improvement over 0-shot evaluation, suggesting the critical role the retrieval-based approach might play in enhancing OpenFlamingo's performance.

We further prompted LLaVa-13B and mPlug-owl-7B by augmenting the questions prompts with the data tables extracted by the DePlot model (see Section~\ref{chartvqaexpertsystems}). LLaVa-13B received a considerable performance boost compared to mPLug-owl-7B, indicating that a sufficiently large LLM backbone may better utilize the data-table strings. In another experiment, we prompted GPT-3.5 with the same augmented questions prompts (denoted as DePlot + GPT-3.5 in the Table~\ref{tab:evaluation_large_multimodal_models_performance}) and observed GPT-3.5 beats the majority of the baselines except LLaVA-13B. This finding highlights the importance of an expert chart-to-table augmentation technique.


Lastly, we present the how the best performing model, LLaVa-13B, improve after being fine-tuned with the SciGraphQA dataset. We refer to this fine-tuned model as SciGrahQA-baseline, and as seen in Table~\ref{tab:evaluation_large_multimodal_models_performance}, outperforms the same model without fine-tuning (CIDEr score of 0.268 vs 0.153). As shown in the experiments on effect of training set size in the following section, this baseline model may be constrained by the number of trainable parameters and the linear projector between image tokens and language tokens in the LLaVA architecture. Within the scope of this paper, we did not perform model architecture search or hyperparameter tuning. 


\begin{figure}[!htbp]
\centering
\includegraphics[width=8.74cm,height=6.31cm]{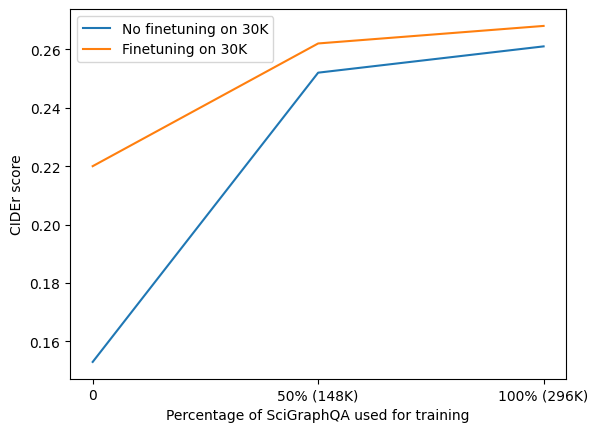}
\caption{Impact of the dataset size and training with prompt augmentation on fine-tuning performance. For the latter, we constructed 30K DePlot-enhanced SciGraphQA samples. The LLaVA-13B checkpoint was used as the base model in our fine-tuning experiments.}
\end{figure}


We also studied how the size of the training set affects the fine-tuning performance. We found a positive correlation between the dataset size and model performance, with the most significant improvement seen when using the first 50\% of the dataset, although training on the full dataset also provided benefits. The lesser performance gain from 50\% to 100\% of the dataset may be attributed to the utilization of the LoRa in the LLM backbone, rather than full fine-tuning as done in \cite{liu2023}. We found that during the pre-training of the LLaVA-style linear projector, the loss plateaued early, suggesting further enhancements could be achieved by increasing trainable parameters in LoRa or using a more powerful multi-modal adapter. Furthermore, as indicated in Section~\ref{finetuningmethod}, we further fine-tuned the model with a 30K subset of the dataset augmented with the data tables. As shown in Figure~\ref{tab:abiliation_experiment_performance_different_setup}, fine-tuning with this additional subset consistently enhanced the CIDEr score across the board.

\section{Discussion}

We introduced SciGraphQA, a dataset of 300K multi-turn question-answering dialogues derived from actual academic publications. This dataset is distinguished by its focus on scientific graphs, its large size (13 times larger than ChartQA), and the comprehensive textual context provided for each dialogue (including the paper's title, abstract, and the paragraph containing the graph). The quality of the responses in SciGraphQA was verified by an impressive 8.7/10 GPT-4 average rating on a test set. We believe that this dataset will be a valuable resource for the development of multimodal large language models.

From a modeling perspective, we employed a fixed CLIP/ViT vision encoder for our 0-shot evaluations and fine-tuning results, with mPLUG-owl being the only exception. CLIP had been extensively pre-trained on a massive dataset of web-sourced images, so enabling the encoder to adapt during training could have potentially improved the integration of visual elements into the LLM input domain, especially for graphs. For example, spatial awareness may be necessary to identify specific legends, trends, or labels in order to answer compositional questions like "where do the blue and green lines intersect?". However, as CLIP was trained to align visual embeddings with captions, its spatial awareness abilities may be limited. We hypothesize that segmentation-based approaches, such as employing the Segment-Anything Model \cite{kirillov2023segment}, to extract masks and project them into the LLM token embedding space, could further improve the visual-question-answering capability of MLLMs.

In this study, we confined our evaluations to conventional NLP metrics, which only take into account pairs of ground truth reference answers and model-generated predictions. However, GPT-4 could be harnessed as an evaluation tool due to its capability in evaluating the model's predicted answers based on the entire text context in conjunction with the posed questions. In our experimentation with GPT-4, however, we encountered three notable challenges.
First, inconsistent results were observed when asking for a rating directly and when soliciting "step-by-step" thinking in the Chain-of-thought prompting style. Second, limited resources constrained us to run only 100 visual question-answering prompts for GPT-4 evaluations (for context, the authors of LLaVa used only 30 samples for their quantitative evaluation \cite{liu2023}. This hindered conclusive comparisons between the models.
Third, the non-deterministic property of GPT-4 evaluation could render subsequent work on SciGraphQA incomparable with our published results. Therefore, we reported only the CIDEr, ROUGE, and BLEU-4 scores, which are freely accessible, open, and reproducible. We would like to stress that further exploration of GPT-4 as an evaluation tool is a worthwhile endeavor and could potentially be integrated as a reward model for reinforcement learning.

Adapting MLLMs to handle out-of-distribution data, particularly in scientific graph question-answering is a significant challenge given the costly and complex nature of fine-tuning large foundation models, their sensitivity to hyperparameters and input data quality, and more importantly the risk of catastrophic forgetting. We explored the use of LoRA to mitigate the fine-tuning cost and further used prompt augmentation to alleviate the remaining issues.

\section{Conclusion}
We introduced SciGraphQA, a novel large-scale dataset for multi-turn question answering on scientific graphs. With 295K samples derived from over 290,000 academic papers, SciGraphQA represents the largest open-sourced dataset in this domain with real-world, non-synthetic graphs and with open-ended conversations not bound by fixed templates of questions or vocabulary. The multi-turn dialogues of SciGraphQA were generated by prompting a commercial LLM using comprehensive textual context including titles, abstracts, captions, and pertinent paragraphs. This extensive context enables dialogues that capture the nuanced process of explaining complex graphs, unlike most existing VQA datasets focused on simple facts or graph composition questions. Looking ahead, SciGraphQA provides a valuable resource for pre-training and instruction tuning of MLLMs to handle real-world scientific graphs. By releasing SciGraphQA as an open dataset, we aim to spur progress on this frontier.



\section{References}

\bibliographystyle{unsrtnat}
\bibliography{SciGraphQA_3.bib}
\section{Appendix}\label{appendix}
\textbf{Palm-2-chat prompt}

 You are an AI visual assistant that can analyze a graph in a scientific paper. You are provided with the OCR-extracted text, the caption of the figure, and the first paragraph that mentioned the figure.

 Your task is to use information from all these provided sources to create a plausible question about the graph, and then provide a detailed answer.

 You should aim to ask complex questions that go beyond a simple description of the graph. The answer to such questions should require understanding the graph data, and then reasoning based on background knowledge or interpretation. You should aim to provide guides and scholarly perspectives based on the graph's data and context.

 Avoid directly revealing specific details in the question. Make the question challenging in a way that the user needs to first reason about the graph data and the context they have derived from the accompanying paper.

 Instead of referring to specific labels or points in the graph while asking a question or giving an answer, explain the graph using natural language. Include details like data trends, prominent points, and relationships between different data sets visualized in the graph.

 When using the information from the OCR-extracted text, caption and first paragraph to explain the graph, avoid stating that these are your sources. Always answer as if you are directly interpreting the graph according to your AI comprehension, understanding and reasoning. Regarding the format of the given input: the following context described the image:  "from": "human", "value": contains OCR extracted text listed, and 'Figure (integer) caption of the image'."from: gpt value: contains the first paragraph in the text that mentioned the figure.

 Examples:

 Example 1:

 User: The following context described the image:  "from": "human", "value": contains OCR extracted text listed, and 'Figure (integer) caption of the image'."from: gpt value: contains the first paragraph in the text that mentioned the figure.

 Assistant: Question: How does the percentage of groups that have successfully acquired a timetable change along with fluctuations in the group size, and based on the graph, which areas are more likely to have successful timetable allocation?

 Answer: The graph presents a relationship between group size and the success rate in acquiring a timetable. It highlights a trend where as group size increases, the likelihood of acquiring a timetable seems to diminish. This is the overall pattern across all the regions, including Scotland, Central UK, and South and Central UK.

 Example 2:

 User：

The following context described the image: "from": "human", "value": "OCR extracted text list, separated by ', ' : SNR $=$ 20DB0.5INR (dB)(a)SNR $=$ 40dB1050INŘ (dB)(b)SNR $=$ 40dB1.50.5INR (dB)(c)gap (bits/sec/Hz)gap (bits/sec/Hz)gap (bits/sec/Hz)$\%$24, SNR, $=$, 20DB, 0.5, INR, (dB), (a), SNR, $=$, 40dB, 10, 50, INŘ, (dB), (b), SNR, $=$, 40dB, 1.5, 0.5, INR, (dB), (c), gap, (bits/sec/Hz), gap, (bits/sec/Hz), gap, (bits/sec/Hz), $\%$24 Fig. 7. Numerical analysis: gap between achievable scheme and the outerbounds in Corollary 1 at (a) SNR $=$ 20dB, (b) SNR $=$ 40dB, and (c) SNR $=$ 60dB for CFB $=$ 10 bits.<image> Give an elaborate explanation of the image you see" $\}$, $\{$ "from": "gpt", "value": "(i) k : k $\in$ $\{$1, 2$\}$, i $\in$ $\{$1, 2, 3, 4$\}$). In general, one can solve an optimization problem to find the optimal choice of power level assignments that maximizes the achievable sum-rate. We have performed numerical analysis for this optimization problem. Figure 7 shows the gap between our proposed achievable scheme and the outer-bounds in Corollary 1 at (a) SNR $=$ 20dB, (b) SNR $=$ 40dB, and (c) SNR $=$ 60dB, for CFB $=$ 10 bits. In fact through our numerical analysis, we can see that the gap is at most 4, 5, and 5.5 bits/sec/Hz for the given values of SNR, respectively. Note that sharp points in Figure 7 are due to the change of achievability scheme for different values of INR as described before." $\}$

 Answer:

 Question: What is the relationship between the gap between the achievable scheme and the outer-bounds in Corollary 1, and the SNR and INR values?

 Answer: The graph shows the gap between the achievable scheme and the outer-bounds in Corollary 1 at (a) SNR $=$ 20dB, (b) SNR $=$ 40dB, and (c) SNR $=$ 60dB, for CFB $=$ 10 bits. The gap is at most 4, 5, and 5.5 bits/sec/Hz for the given values of SNR, respectively. The sharp points in the graph are due to the change of achievability scheme for different values of INR.

 Example 3:

 User [insert prompt regarding current image]
"""
\end{document}